\documentclass[10pt,twocolumn,letterpaper]{article}

\usepackage[pagenumbers]{cvpr} %

\usepackage[dvipsnames]{xcolor}

\definecolor{cvprblue}{rgb}{0.21,0.49,0.74}
\usepackage[pagebackref,breaklinks,colorlinks,citecolor=cvprblue,linkbordercolor=cvprblue,urlcolor=cvprblue,hyperindex=true]{hyperref}

\usepackage[disable]{todonotes}

\usepackage{xspace}
\usepackage[utf8]{inputenc} %
\usepackage[T1]{fontenc}    %
\usepackage{url}            %
\usepackage{booktabs}       %
\usepackage{amsfonts}       %
\usepackage{nicefrac}       %
\usepackage{microtype}      %
\usepackage{array}
\usepackage{wrapfig}
\usepackage[font=footnotesize,labelfont=bf]{caption}
\usepackage{duckuments}

\usepackage{tcolorbox}

\usepackage{times}
\usepackage{epsfig}
\usepackage{graphicx}
\usepackage{amsmath}
\usepackage{amssymb}
\usepackage[dvipsnames]{xcolor}
\usepackage{booktabs,arydshln}
\usepackage{xspace}
\usepackage{multirow}
\usepackage[normalem]{ulem}
\usepackage{duckuments}
\usepackage{bbm}
\usepackage{tabularx,ragged2e}
\usepackage{pifont}
\newcommand{\cmark}{\ding{51}}%
\newcommand{\xmark}{\ding{55}}%
\newcolumntype{Y}{>{\RaggedRight\arraybackslash}X}

\usepackage{enumitem}

\definecolor{ForestGreen}{RGB}{34,139,34}

\usepackage{algpseudocode}
\usepackage[font=small,labelfont=bf]{caption}
\usepackage[rightcaption]{sidecap}    %
\sidecaptionvpos{figure}{c}

\usepackage{array}
\usepackage{multirow}
\usepackage{booktabs}
\usepackage[linesnumbered,ruled,vlined]{algorithm2e}
\usepackage[font=scriptsize,labelfont=bf]{subcaption}
\usepackage[normalem]{ulem}
\usepackage{xparse}
\usepackage{pifont}
\usepackage{bm}
\usepackage{epigraph}

\usepackage{tabularx}
\usepackage{ifthen}
\usepackage{makecell}

\usepackage{wrapfig}

\usepackage[normalem]{ulem}

\usepackage[scaled=0.85]{DejaVuSansMono}

\SetCommentSty{mycommfont}
\SetKwInput{KwInput}{Input}                %

\usepackage{microtype}

\setlist{nosep} %

\usepackage{listings}

\definecolor{codegreen}{rgb}{0,0.6,0}
\definecolor{codegray}{rgb}{0.5,0.5,0.5}
\definecolor{codepurple}{rgb}{0.58,0,0.82}
\definecolor{backcolour}{rgb}{1.,1.,1.}

\lstdefinestyle{mystyle}{
    backgroundcolor=\color{backcolour},
    commentstyle=\color{codegreen},
    keywordstyle=\color{magenta},
    numberstyle=\tiny\color{codegray},
    stringstyle=\color{codepurple},
    basicstyle=\ttfamily\footnotesize,
    breakatwhitespace=false,
    breaklines=true,
    captionpos=b,
    keepspaces=true,
    numbers=left,
    numbersep=5pt,
    showspaces=false,
    showstringspaces=false,
    showtabs=false,
    tabsize=2
}

\usepackage[title]{appendix}

\makeatletter
\newcommand{\printfnsymbol}[1]{%
    \textsuperscript{\@fnsymbol{#1}}%
}
\makeatother

\def\modelname{BootPIG\xspace}

\title{BootPIG: Bootstrapping Zero-shot Personalized Image Generation Capabilities in Pretrained Diffusion Models}

\author{
Senthil Purushwalkam\thanks{Equal contribution.}
\and
Akash Gokul\footnotemark[1]
\and
Shafiq Joty
\and
Nikhil Naik\vspace{3pt}
\\
Salesforce AI Research\\
{\tt\small \{spurushwalkam,agokul,sjoty,nnaik\}@salesforce.com} 
}

\begin{document}

\twocolumn[{%
\renewcommand\twocolumn[1][]{#1}%
\maketitle
\begin{center}
    \centering
    \captionsetup{type=figure}
    \vspace{-20pt}
    \includegraphics[width=0.73\textwidth]{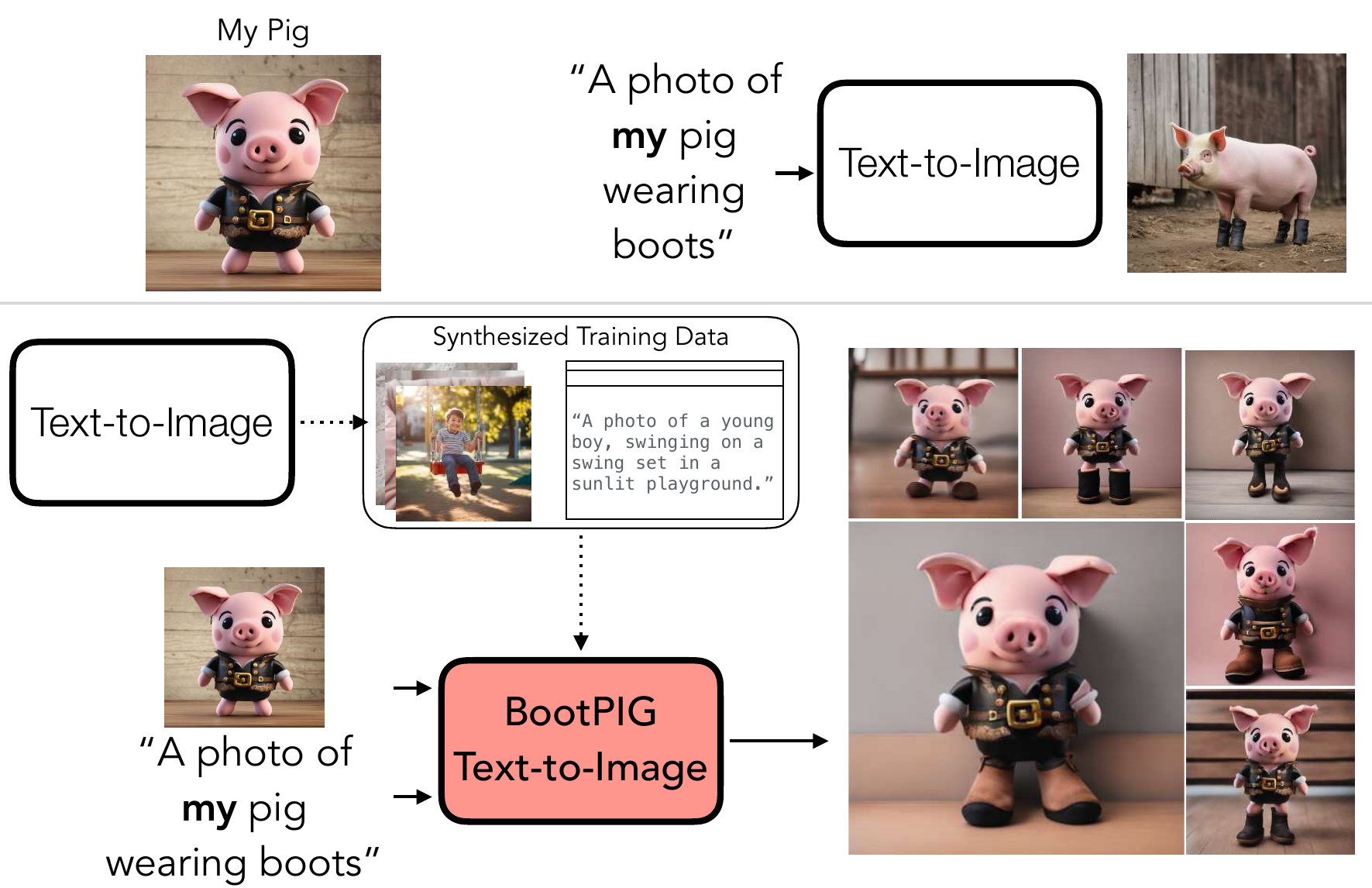}
    \captionof{figure}{Existing text-to-image models demonstrate exceptional image synthesis capabilities. However, they  fail to "personalize" generations according to a specific subject. \modelname (Ours) enables zero-shot subject-driven generation through a \emph{bootstrapped} training process that uses images synthesized by the text-to-image model (Bottom). \modelname trained text-to-image models can synthesize novel scenes containing the input subject \emph{without test-time finetuning} while maintaining high fidelity to the prompt and subject.}
    \label{fig:fig0}
\end{center}%
}]

\begin{abstract}
\vspace{-8pt}
    Recent text-to-image generation models have demonstrated incredible success in generating images that faithfully follow input prompts. However, the requirement of using words to describe a desired concept provides limited control over the appearance of the generated concepts. In this work, we address this shortcoming by proposing an approach to enable personalization capabilities in existing text-to-image diffusion models. We propose a novel architecture (BootPIG) that allows a user to provide reference images of an object in order to guide the appearance of a concept in the generated images.

    The proposed BootPIG architecture makes minimal modifications to a pretrained text-to-image diffusion model and utilizes a separate UNet model to steer the generations toward the desired appearance. We introduce a training procedure that allows us to bootstrap personalization capabilities in the BootPIG architecture using data generated from pretrained text-to-image models, LLM chat agents, and image segmentation models. In contrast to existing methods that require several days of pretraining, the BootPIG architecture can be trained in approximately 1 hour. Experiments on the DreamBooth dataset demonstrate that BootPIG outperforms existing zero-shot methods while being comparable with test-time finetuning approaches. Through a user study, we validate the preference for BootPIG generations over existing methods both in maintaining fidelity to the reference object's appearance and aligning with textual prompts.
\end{abstract}

\section{Introduction}
\label{sec:intro}

Over the last few years, research in generative modeling has pushed the boundaries of technological innovation and enabled new avenues of creative expression. In computer vision, text-to-image generation models \cite{ramesh2021zero, ding2022cogview2, ramesh2022hierarchical, rombach2022high, chang2023muse, yu2023scaling, yu2022scaling, kang2023scaling} have showcased remarkable proficiency in generating high-fidelity images corresponding to novel captions. However, the ability to generate arbitrary images depicting a chosen caption has limited applications. Recent advancements in research have expanded upon these text-to-image models, leveraging their capabilities to solve more widely applicable tasks such as image inpainting\cite{xie2023smartbrush}, image editing\cite{wallace2023edict,meng2021sdedit}, style transfer\cite{zhang2023inversion}, and even generation of 3D models from text\cite{poole2022dreamfusion} and images\cite{melas2023realfusion, xu2023neurallift, purushwalkam2023conrad}. One such task with numerous applications is the problem of \textit{Personalized Image Generation}.

Personalized Image Generation (also know as subject-driven generation) is the ability to generate images of specific \textit{personal} objects in various user-specified contexts. For instance, one may want to visualize what their pet would look like if it wore cowboy boots. Beyond such personal visualization experiments, this capability has the potential to serve as a versatile tool with applications ranging from personalized storytelling to interactive design. 

More concretely, the Personalized Image Generation problem can be formulated as follows: given a few reference images depicting an object and a target caption, the goal is to generate an image that corresponds to the caption while faithfully capturing the object's appearance. In recent research, several methods have been developed to leverage pretrained text-to-image models to accomplish this task. These approaches can be divided into two categories-- Test-time Finetuning and Zero-Shot Inference methods. The former category involves methods that finetune the text-to-image model parameters or learn additional parameters to learn the subject's appearancee. These methods often take a few minutes to a few hours to update parameters before personalized images are generated. The latter category includes methods that rely on pretraining the text-to-image models and do not involve test-time updates. While Zero-Shot Inference-based methods are more user-friendly and efficient for deployment, they fall short compared to Test-time Finetuning methods in terms of faithfulness to the reference object and textual prompt following capabilities.

In this work, we propose a novel solution to the Personalized Image Generation problem that provides the efficiency of Zero-Shot Inference methods while outperforming existing Test-time Finetuning methods. We introduce a novel architecture, named \modelname, designed to enable personalized image generation capabilities in a pretrained text-to-image generation model. The \modelname architecture comprises two replicas of the pretrained text-to-image model — one dedicated to extracting visual features from reference images and another for the actual image generation process.

In order to train our proposed model, we present a novel procedure that does not directly utilize any human curated dataset. Instead, we bootstrap the personalization capability by learning from data synthetically generated using pretrained text-to-image generation models, state-of-the-art chat agents and image segmentation models. Furthermore, unlike existing Zero-Shot Inference methods that require several days of compute for pretraining, our proposed model can be trained in approximately 1 hour on 16 A100 GPUs.

Experiments on the DreamBooth dataset show that \modelname generations outperform existing zero-shot methods while performing comparably to test-time finetuned approaches based on existing metrics. After conducting a user study, which asks users to compare generations from two approaches based on subject fidelity and prompt fidelity, we find that users generally prefer \modelname generations over existing zero-shot and test-time finetuned methods. 

The contributions of our work are summarized as follows:
\begin{itemize}
  \item We propose a novel architecture that enables zero-shot subject-driven generation while minimally modifying the architecture of the pretrained text-to-image model and requiring only 1 hour to train.  
  \item We demonstrate an effective bootstrapped learning procedure which does not require human-curated data and allows a pretrained text-to-image to use its own features to learn subject-driven generation.
  \item \modelname excels in zero-shot personalized image generation outperforming existing zero-shot method and test-time finetuned methods based on quantitative evaluations and user studies.
\end{itemize}

\section{Related Work}
\label{sec:related}

\subsection{Text-to-Image Synthesis}
Progress in generative models has led to breakthroughs in text-to-image synthesis. Existing text-to-image models \cite{ramesh2021zero, ding2022cogview2, ramesh2022hierarchical, rombach2022high, chang2023muse, yu2023scaling, yu2022scaling, kang2023scaling} are capable of generating high-quality images in accordance with a given text prompt. These models fall into one of the following categories: diffusion models \citep{ramesh2022hierarchical, saharia2022photorealistic, rombach2022high}, autoregressive image models \citep{ramesh2021zero, yu2022scaling, yu2023scaling}, non-autoregressive image models \cite{chang2022maskgit}\citep{chang2023muse}, and generative adversarial networks (GANs) \citep{goodfellow2020generative}\citep{kang2023scaling}. While these models demonstrate exceptional image generation and fidelity to an input prompt, their outputs are constrained by the limitations of the text interface. This hinders their ability to generate images with a specific subject or follow additional controls from signals such as images. In this work, we extend the capability of pretrained text-to-image diffusion models to enable zero-shot subject-driven generation by introducing subject images as an additional source of control.

\subsection{Subject-Driven Text-to-Image Synthesis via Test-time Finetuning}
Following the breakthroughs in text-to-image generation, many works extended these models to enable subject-driven generation. Subject-driven generation (personalized image generation) focuses on synthesizing an input subject in novel scenes. The seminal works of Textual Inversion \citep{gal2022image} and DreamBooth \citep{ruiz2023dreambooth} introduced test-time finetuning approaches that customized a text-to-image generative model given a few images of a subject. Textual Inversion enables subject-driven generation by learning a unique text token for each subject. In contrast, DreamBooth finetunes the entire text-to-image backbone and demonstrates superior results in fidelity to the subject. Subsequent works \citep{kumari2023multi, hao2023vico, voynov2023p+, gal2023encoder, arar2023domain, gu2023mix, smith2023continual, ruiz2023hyperdreambooth} extended these approaches by developing improved finetuning methods. In particular, CustomDiffusion \citep{kumari2023multi} and ViCo \citep{hao2023vico} improve subject-driven generation by combining ideas from Textual Inversion and DreamBooth, by learning a specific text embedding for each concept while finetuning the cross-attention layers of the text-to-image model. Unlike these works, our method enables zero-shot subject-driven generation and avoids the need to train unique models for each subject.

\subsection{Zero-Shot Subject-Driven Text-to-Image Synthesis}
Subject-driven generation approaches, such as DreamBooth and Textual Inversion, require hundreds, sometimes even thousands, of steps to learn the user-provided subject. Recent works have sought to avoid this tedious finetuning process by developing zero-shot methods for subject-driven generation. These methods typically pretrain image encoders across large datasets to learn image features that improve the generative model's ability to render the subject. InstantBooth \cite{shi2023instantbooth} introduces adapter layers in the text-to-image model and pretrains an adapter for each concept, \eg cats. In comparison, our method is not restricted to learning single categories and, instead, can generalize to novel concepts. The authors of \citet{jia2023taming} propose to use frozen CLIP \citep{radford2021learning} image encoders and continue training the generative backbone to use CLIP image features. Similar to this work, our approach continues training the generative backbone. However, our approach does not introduce additional image encoders and does not require real data. Along the lines of pretraining image encoders, BLIP-Diffusion \cite{li2023blipD} pretrains a Q-Former \citep{li2023blip} that learns image features which are aligned with text embeddings. Similarly, ELITE \cite{wei2023elite} pretrains two image encoders– one that learns image features that align with token embeddings, and a second encoder that learns image features that align with the text encoder’s features. In comparison, our approach avoids introducing a bottleneck in the form of text-aligned features and, instead, uses the features from the generative backbone to guide generation.

\subsection{Multimodal Controls for Text-to-Image Models}
Concurrent to the work in subject-driven generation, many works have explored extending the conditional controls in text-to-image models. These methods introduce learnable components, such as modality-specific adapters \citep{li2023gligen, zhao2023uni, mou2023t2i, hu2023cocktail, kim2023diffblender, ye2023ip}, to enable additional control via inputs such as depth maps or color palettes. In particular, ControlNet \citep{zhang2023adding} demonstrated exceptional results in controllable generation by training a learnable copy of the generative model’s backbone to accept new input conditions. Our architecture shares similarities with the ControlNet architecture, as both works use a copy of the generative backbone. However unlike ControlNet, our work focuses on zero-shot subject-driven generation.

\section{Method}
\label{sec:approach}

In this section, we present the details of our proposed approach for enabling personalized image generation capabilities in existing latent diffusion models. Given a caption and a set of reference images that depict an object, the goal is to generate an image that follows the caption while ensuring that the appearance of the object matches that of the reference images. We accomplish this by introducing a novel architecture (\modelname) that builds on top of existing diffusion models for text-to-image generation.

\subsection{Preliminary: Diffusion Models}
\label{subsec:diffusion_intro}
Diffusion models \citep{sohl2015deep, song2019generative, ho2020denoising} learn a data distribution by iteratively denoising samples from a Gaussian distribution. Given a data sample $x$, a noised version $x_t := \alpha_{t}x + \sigma_{t}\epsilon$, is generated by applying noise, $\epsilon \sim \mathcal{N}(0, 1)$ according to a timestep $ t \in \{1, \ldots, T\}$, where $T$ is the total number of timesteps, and $\alpha_{t}$, $\sigma_{t}$ control the noise schedule. The model, $\epsilon_\theta$, is trained using the mean squared error objective between the noise $\epsilon$ and the predicted noise $\epsilon_{\theta}(x_t, t, c)$, where $c$ refers to a conditioning vector \eg a text prompt (Equation \ref{eq:diffusion_loss}).

\begin{equation}
    E_{x, c, \epsilon, t \sim \mathcal{U}([0, T])} \left\| \epsilon - \epsilon_{\theta}(x_t, t, c) \right\|_{2}^{2}
    \label{eq:diffusion_loss}
\end{equation}

In this work, we use pretrained text-to-image diffusion models and do not alter the diffusion objective during our training. Instead, we introduce additional image features into the architecture which are trained using Equation \ref{eq:diffusion_loss}.

\subsection{\modelname Model Architecture}
\label{subsec:model}

\begin{figure}
    \includegraphics[width=\columnwidth]{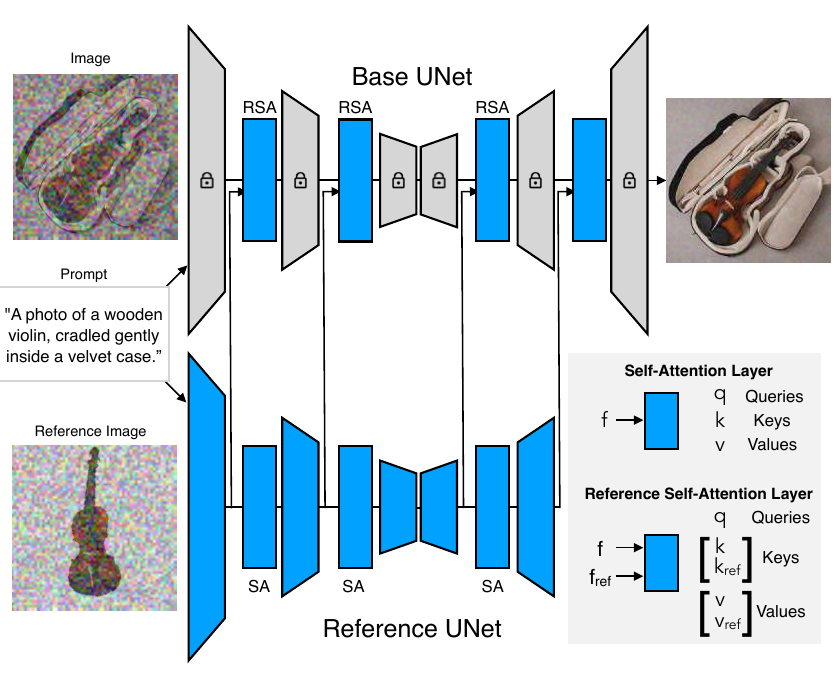}
    \caption{\textbf{Model Architecture}: We propose a novel architecture, that we refer to as \modelname, for personalized image generation. The model comprises of two replicas of a latent diffusion model - Reference UNet and Base UNet. The Reference UNet processes reference images to collect the features before each Self-Attention (SA) layer. The Base UNet's SA layers are modified to Reference Self-Attention (RSA) layers that allow conditioning on extra features. Using the collected reference features as input, the Base UNet equipped with the RSA layers estimates the noise in the input to guide the image generation towards the reference objects.}
    \label{fig:architecture}
\end{figure}

We present an overview of the \modelname architecture in Figure~\ref{fig:architecture}. The key idea of our proposed architecture is to inject the appearance of a reference object into the features of a pretrained text conditioned image diffusion model, such that the generated images imitate the reference object. In this work, we use Stable Diffusion~\cite{rombach2022high} as our pretrained text-to-image diffusion model. Stable Diffusion is a Latent Diffusion Model \cite{rombach2022high} that uses a U-Net \cite{ronneberger2015u} architecture consisting of Transformer\cite{vaswani2017attention} and Residual\cite{he2016deep} blocks. Our proposed \modelname architecture modifies the information processed by the self-attention layers in the Transformer blocks in order to control the appearance of the generated objects. Let the Stable Diffusion U-Net model be denoted by $U_\theta(x, c, t)$ where $x_t$ are noisy input latents, $c$ is an input textual prompt and $t$ is the timestep in the diffusion process.

\noindent
\textbf{Injecting Reference Features}~~~~
A self-attention (SA) layer that receives a latent feature $f\in \mathbb{R}^{n\times d}$, performs the following operation:

\begin{align}
    \texttt{SA}(f) = W_o ~ \Big( \texttt{softmax}\Big( ~ q(f) k(f)^T ~ \Big) v(f) \Big)
\end{align}
where $q, k, v$ are linear mappings known as the query, key and value functions with parameters $W_q, W_k, W_v \in \mathbb{R}^{d\times d'}$ respectively that project the features to a chosen dimension $d'$. $W_o \in \mathbb{R}^{d'xd}$ projects the output back to the original dimension $d$. We propose to replace all the Self-Attention (SA) layers with an operation that we refer to as Reference Self-Attention (RSA), that allows us to inject reference features. The RSA operator takes as input the latent features $f\in \mathbb{R}^{n\times d}$ and reference features of the same dimension $f_{\texttt{ref}}\in \mathbb{R}^{n_\texttt{ref}\times d}$, and performs the following operation:

\begin{align}
    \resizebox{0.85\columnwidth}{!}{$
            \texttt{RSA}(f, f_\texttt{ref}) = W_o ~ \Big(\texttt{softmax}\Big( ~ q(f) \begin{bmatrix} k(f) \\ k(f_\texttt{ref})\end{bmatrix}^T  \begin{bmatrix} v(f) \Big] \\ v(f_\texttt{ref})\end{bmatrix} \Big)$
    }
\end{align}

where $[ : ]$ indicates concatenation along the first dimension. Intuitively, the RSA operator facilitates the injection of reference features, allowing the diffusion model to ``attend'' to them in the computation of the output latent feature. Let us denote this new U-Net, referred to as Base U-Net, by $U^\texttt{RSA}_\theta(x, c, t, \{f_\texttt{ref}^{(1)}, f_\texttt{ref}^{(2)},...,f_\texttt{ref}^{(L)}\})$ containing $L$ RSA layers. For simplicity, we use the notation $\mathcal{F}_\texttt{ref}$ to denote the set of $L$ reference features. Note that the RSA operator does not introduce any new parameters and reuses the weight parameters $W_o, W_q, W_k, W_v$.

\noindent
\textbf{Extracting Reference Features}~~~~ Given a reference image $\mathcal{I}^\texttt{ref}$, we need to extract the appropriate features $f_\texttt{ref}$ that can be passed to each RSA layer. We propose to extract features using a separate U-Net $U_\phi(x, c, t)$, referred to as Reference U-Net, that follows the same architecture as the Base U-Net and is initialized with the same parameters ($\phi=\theta$). For a given $t$, we perform the forward diffusion process on the reference image $\mathcal{I}_\texttt{ref}$ to compute the noised reference latents ${x'_\texttt{ref}}_t$. We pass ${x'_\texttt{ref}}_t$ as input along with the textual prompt, and extract the features before the $L-$ SA layers as $\mathcal{F}_\texttt{ref}$. This ensures that the extracted reference features have the appropriate dimensions and are compatible with the weights of the RSA layers.

\subsection{Training}
\label{subsec:training}

The \modelname architecture allows us to pass features of a reference image to the RSA layers. However, since the original diffusion model $U_\theta$ was not trained with RSA layers, we observe that the generated images are corrupted (see supplementary material) and do not correctly follow the input prompts. In order to rectify this, we propose to finetune the parameters of the Reference U-Net $\phi$ in order to extract better reference features and the parameters of the RSA layers ($W_o, W_q, W_k, W_v$) to better utilize the reference features.

Given a dataset of triplets containing a reference image, a textual prompt and a target image following the textual prompt while accurately depicting the reference object, we finetune the \modelname architecture using the same objective as the original latent diffusion model (see Section~\ref{subsec:diffusion_intro}). The Reference U-Net takes as input noisy VAE latents (noised according to timestep $t$) corresponding to the reference image as input, along with the timestep $t$ and the target caption. The Base U-Net  receives as input noisy VAE latents corresponding to the target image (similarly noised), the timestep $t$, the target caption and the reference features collected from the Reference U-Net. The parameters of the Reference U-Net and the RSA layers are updated to accurately estimate the noise in the input latents (see Eq~\ref{eq:diffusion_loss}). In order to preserve the Base U-Net's prompt following capabilities, we randomly drop the reference image's features (with probability 0.15), thereby reverting the Base U-Net model back to the SA based architecture. We provide a more detailed description of the training pipeline along with pseudocode in the supplementary material.

\begin{figure}
    \includegraphics[width=\columnwidth]{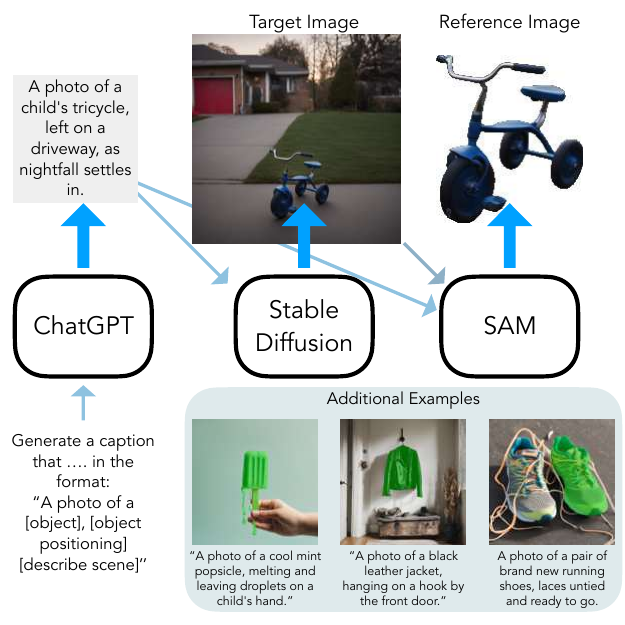}
    \caption{\textbf{Synthetic Training Data:} We propose an automated data generation pipeline to generate (reference image, target image, target caption) triplets for training \modelname. The pipeline uses ChatGPT to generate captions, Stable Diffusion to generate images and the Segment Anything Model to segment the foreground which serves as the reference image.}
    \label{fig:dataset}
\end{figure}

\subsection{Bootstrapping Training Data}
\label{subsec:data}

Collecting a large curated source of training data for optimizing the \modelname architecture is an expensive if not infeasible process. This would involve gathering paired reference and target images depicting the same object instance in different contexts. Instead, we propose a synthetic data generation pipeline that leverages the capabilities of existing pretrained Computer Vision and NLP models.

In Figure~\ref{fig:dataset}, we provide an overview of the data generation pipeline. First, we utilize ChatGPT\cite{Brown2020LanguageMA}, a state-of-the-art conversational agent, to generate captions for potential target images. For each caption, we generate an image using Stable Diffusion\cite{rombach2022high}. We then use Segment Anything Model (SAM)\cite{Kirillov2023SegmentA}, a state-of-the-art segmentation model, to extract a foreground mask corresponding to the main object in the caption. We treat the Stable Diffusion generated image as the target image, the foreground object pasted on a white background as the reference image and the ChatGPT generated caption as the textual prompt. While the reference image does not depict the object in an entirely different context, we observe empirically that this synthetic data is sufficient to learn personalization capabilities.

\subsection{Inference}
\label{subsec:inference}
During inference, at each timestep,  we extract reference features $\mathcal{F}_\texttt{ref}$ by passing a noised reference image, the target caption $c$ and the timestep $t$ to the Reference UNet. We use a classifier free guidance \cite{Ho2022ClassifierFreeDG} strategy to estimate the noise in the noisy generation latents $x'_t$ using the Base UNet as:
\begin{equation}
    \small
    \begin{split}
        \epsilon = U^\texttt{SA}_\theta(x'_t, t) + \eta_{im} (U^\texttt{RSA}_\theta(x'_t, t, f_\texttt{ref}) - U^\texttt{SA}_\theta(x'_t, t)) \\
        + \eta_{text+im} * (U^\texttt{RSA}_\theta(x'_t, t, f_\texttt{ref}, c) - U^\texttt{RSA}_\theta(x'_t, t, f_\texttt{ref}, c))
    \end{split}
\end{equation}
where the first U-Net term estimates the noise without any conditioning information, the second U-Net term estimates noise with just image information and the fourth U-Net term uses both images and captions.

The \modelname architecture described so far takes as input a single reference image. This allows us to train the parameters of the model with synthesized (reference, target) image pairs. A common scenario with several applications is the case where multiple reference images are available to us. In order to handle this scenario, we now propose an inference procedure to leverage a trained \modelname model and utilize appearance information from multiple reference images.

Let $f_\texttt{refi}$ be the reference feature for image $i$ at a specific layer. At each RSA layer, we first compute the outputs without any reference feature ($f_\texttt{ref}=f)$ and using each reference feature:
\begin{gather}
    o = RSA(f, f) \\
    o_i = RSA(f, f_\texttt{refi})  ~~~~\forall i\in \{1,2,...,k\}
\end{gather}
For the output generated by each reference $o_i \in \mathbb{R}^{n\times d}$, we compute the pixelwise norm of difference to the reference-less output $o$. We denote this by $n_i = || o_i - o || \in \mathbb{R}^{n\times 1}$. The pixelwise softmax of these norms is then used to weight the contribution of each reference feature. Specifically, we compute the final output as:
\begin{gather}
    p_i = \texttt{softmax} (n_1, n_2, ..., n_k) [i] \in \mathbb{R}^{n\times 1}\\
    o_\texttt{multiref} = o + \sum_i p_i * (o_i - o)
\end{gather}

Intuitively, at each pixel location, we want to use the reference features that make the largest contribution compared to the reference-less output.

\section{Experiments}
\label{sec:experiments}

In this section, we describe the experimental setup and present results of our proposed model, \modelname, for personalized image generation.

\begin{figure*}[h!]
\centering
    \includegraphics[width=0.95\textwidth]{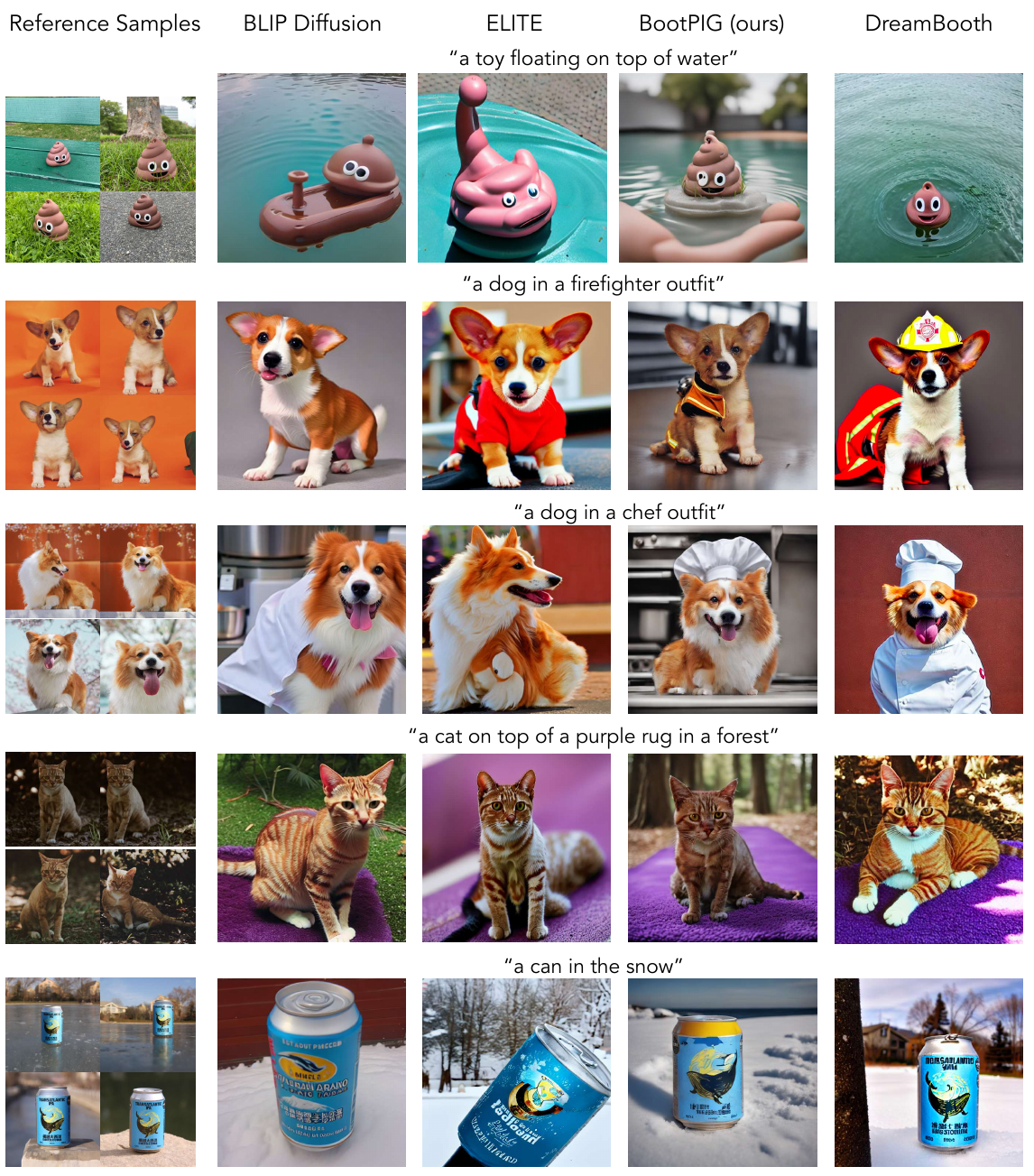}
    \caption{\textbf{Qualitative Comparision}: We provide visual comparisions of subject-driven generations from related methods such as BLIP-Diffusion, ELITE, and DreamBooth. \modelname exhibits high subject and prompt fidelity, outperforming related methods while avoiding test-time finetuning.}
    \label{fig:qualitative-comparision}
\end{figure*}

\subsection{Experimental Setup}
\paragraph{Implementation Details}
The dataset synthesis pipeline discussed in ~\ref{subsec:training} was used to synthesize 200000 (reference image, target image, target caption) triplets as training data. We provide the ChatGPT prompt used to generate these images in the Supplementary. We use the VAE and text encoder weights from the publicly available Stable Diffusion version 2.1 \cite{rombach2022high}. The Base U-Net and Reference U-Net are both initialized from the pretrained Stable Diffusion U-Net weights. During training, we augment the reference images by randomly resizing the foreground, horizontally flipping and moving the foreground to random locations. We train the model using 16 A100 GPUs, with a total batch size of 64 for 10k updates. We use a learning rate of 5e-6 for the Reference U-Net and 1e-6 for the RSA layers of the Base U-Net. The entire training takes 70 minutes, whereas BLIP-Diffusion requires 6 days with the same amount of compute. During inference, we use the UniPC Scheduler\cite{zhao2023uni} and generate images using 50 inference steps. For classifier free guidance (as described in Section~\ref{subsec:inference}), we use $\eta_{im}=5.0$ and $\eta_{text+im}=7.5$ in all the experiments.

\paragraph{Evaluation}
We evaluate our method on the DreamBooth dataset. The dataset contains 30 personalization subjects, each with multiple 3-6 reference images, and 25 novel prompts per subject. Unless specified, we use all available reference images for each generation. We evaluate the zero-shot performance of our method on this dataset using the metrics introduced by the authors of DreamBooth: CLIP-T, CLIP-I, and DINO. CLIP-T is used to measure the alignment between the generated image and the textual prompt. CLIP-I and DINO are used to evaluate the faithfulness of the generated image to the appearance of the reference object.

Additionally, we perform a user-study to compare our method to existing work. We perform a subject fidelity user study and a text alignment user study following the procedure outlined in ELITE \cite{wei2023elite}. In the subject fidelity study, users are provided the original subject image, and two generated images of that subject, generated by different methods (both methods use the same prompt). Users are asked to select which image better represents the original subject. In the text alignment study, users are provided the caption and two generated images, generated by different methods, for that caption and are asked to select which image better represents the caption. 

For comparisons with Textual Inversion and DreamBooth, we use the corresponding implementations in the \texttt{diffusers} package \cite{von_Platen_Diffusers_State-of-the-art_diffusion} to train on each of the 30 subjects. We report CustomDiffusion's evaluation metrics using the results reproduced in \citet{wei2023elite}. For zero-shot methods, such as BLIP-Diffusion and ELITE, we use their official pretrained models to generate images for each subject, following their respective inference procedures.

\subsection{Qualitative Comparisons}
In Figure \ref{fig:qualitative-comparision}, we provide qualitative comparisons between  \modelname and several personalized image generation methods. In comparison to zero-shot methods such as BLIP-Diffusion and ELITE, \modelname exhibits significantly higher fidelity to the original subject. This improved fidelity to the subject is seen at both the object-level and at the level of fine-grained details as seen in examples such as the dog (row 3) or the can (row 5). We attribute this improvement in subject alignment to our reference feature extraction method which, unlike BLIP-Diffusion and ELITE, does not introduce a bottleneck by forcing image features to be aligned with the text features. Compared to DreamBooth, a test-time finetuned approach, we also see improvements in subject fidelity. For example, \modelname better preserves the markings and colors of the cat's fur (row 4) and the facial features, \eg eyes, of the dog (row 3).

\subsection{Quantitative Comparisons}
In Table \ref{tab:quant}, we present the main quantitative comparison of \modelname against existing methods on the DreamBooth dataset. \modelname outperforms all existing zero-shot methods in prompt fidelity ($+1.1$ CLIP-T) and subject fidelity ($+0.8$ CLIP-I, $+2.4$ DINO) metrics. Compared to test-time finetuned methods, \modelname exhibits state-of-the-art performance in prompt fidelity ($+0.6$ CLIP-T) while performing comparably in subject fidelity.

\begin{figure}[h!]
    \includegraphics[width=0.9\columnwidth]{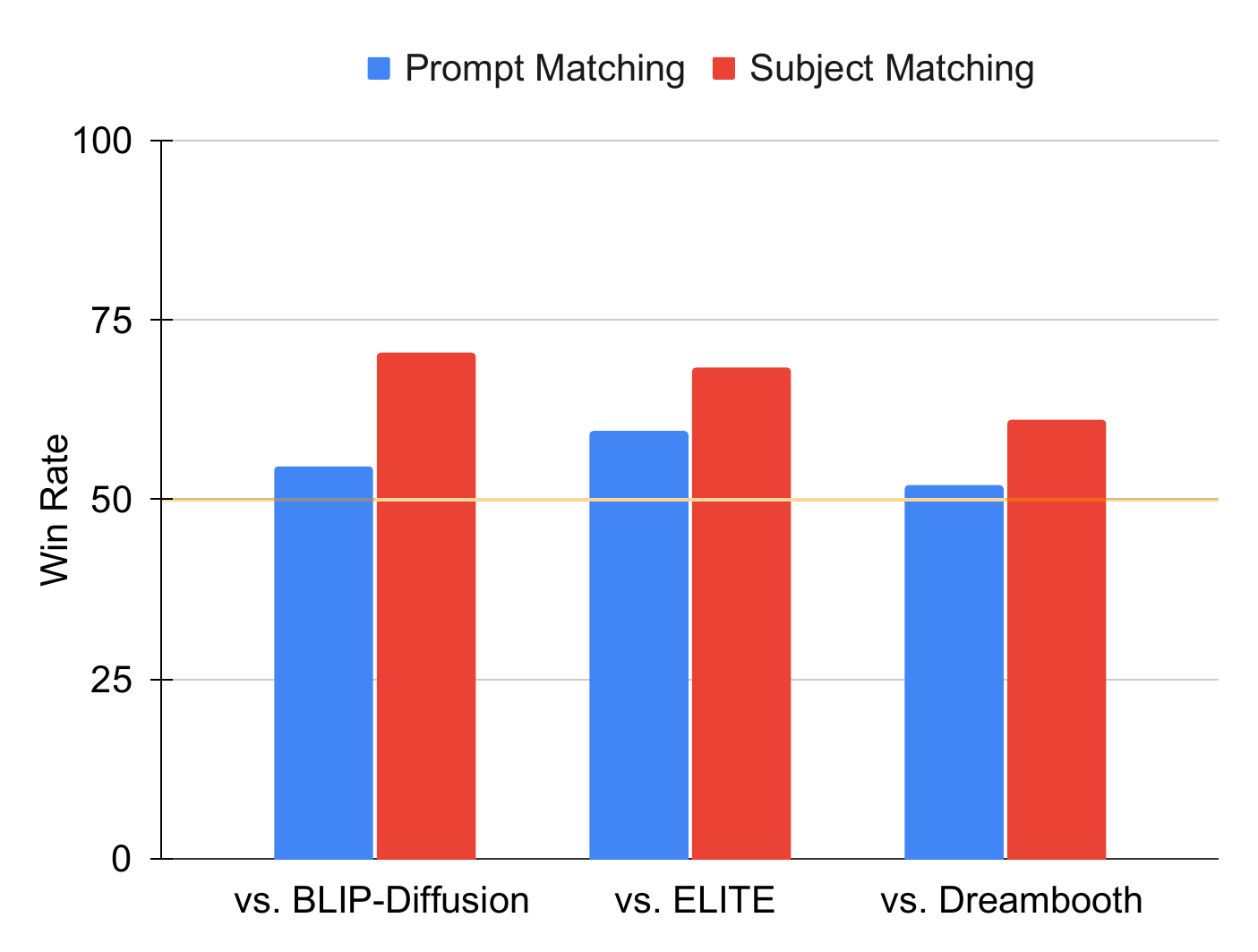}
    \caption{\textbf{User Study}: We report the win rate (\% of users who favored \modelname generations) against existing methods. We perform two studies per head-to-head comparision, one evaluating prompt fidelity and the other evaluating subject fidelity.}
    \label{fig:human-eval}
\end{figure}

\begin{table}[h!]
    \centering
    \caption{\textbf{Quantitative Comparisons:} We report average metrics for subject fidelity (CLIP-I, DINO) and prompt fidelity (CLIP-T) on the DreamBooth dataset.}
    \label{tab:quant}
    \resizebox{0.95\linewidth}{!}{
        \begin{tabular}{@{\extracolsep{\fill}}lcccc@{}}
            \toprule
            Method                                 & Zero-shot & CLIP-T        & CLIP-I        & DINO        \\
            \midrule
            \midrule
            BLIP Diffusion \cite{li2023blipD}      & \cmark    & 30.0          & 77.9          & 59.4          \\
            ELITE \cite{wei2023elite}              & \cmark    & 25.5          & 76.2          & 65.2          \\
            ELITE (reproduced)                     & \cmark    & 29.6          & 78.8          & 61.4          \\
            Ours                                   & \cmark    & \textbf{31.1} & \textbf{79.7} & \textbf{67.4} \\
            \midrule
            Textual Inversion \cite{gal2022image}  & \xmark    & 25.5          & 78.0          & 56.9          \\
            DreamBooth \cite{ruiz2023dreambooth}   & \xmark    & 30.5          & 80.3          & 66.8          \\
            CustomDiffusion \cite{kumari2023multi} & \xmark    & 24.5          & 80.1          & 69.5          \\
            BLIP Diffusion + FT                    & \xmark    & 30.2          & 80.5          & 67.0          \\
            \midrule
            Reference Images                       & -         & -             & 88.5          & 77.4          \\
            \bottomrule
        \end{tabular}
    }
\end{table}

\begin{figure}
    \includegraphics[width=\columnwidth]{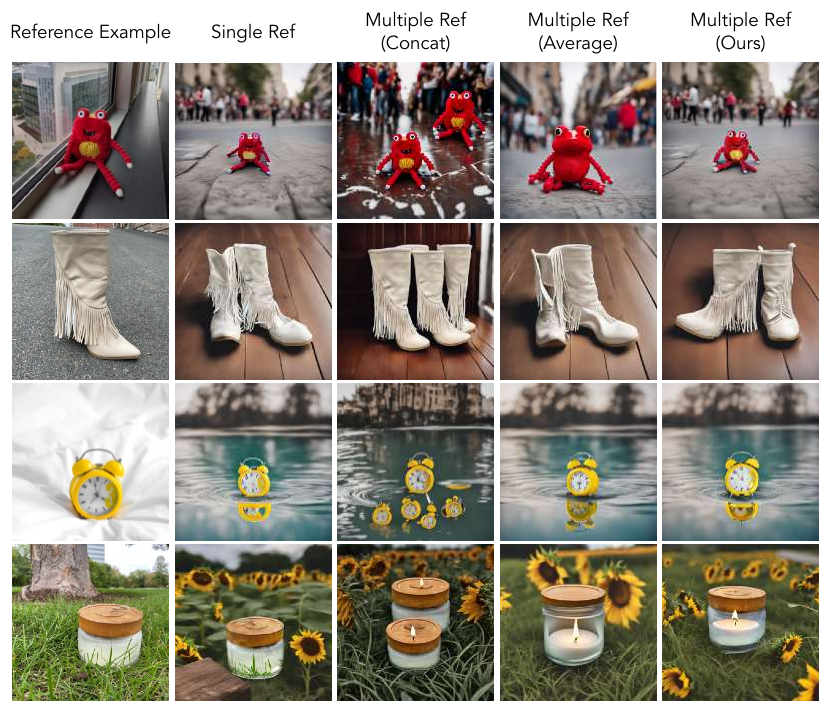}
    \caption{\textbf{Effect of Inference Strategy}: We compare different strategies for pooling information from multiple reference images in the \modelname architecture. Refer to Section \ref{subsec:ablation} for more details.}
    \label{fig:refstrategy}
\end{figure}

From the results of our user study, we find that users consistently prefer \modelname generations over both zero-shot and test-time finetuned methods. Human evaluators find \modelname generations to have significantly greater subject fidelity ($69\%$ win rate versus ELITE $65\%$ win rate versus BLIP-Diffusion, and $62\%$ win rate versus DreamBooth). Additionally, the user study demonstrates that generations from \modelname exhibit higher fidelity to the caption than existing methods ($60\%$ win rate versus ELITE, $54\%$ win rate versus BLIP-Diffusion, and $52\%$ win rate versus DreamBooth). These results underscore the efficacy of our training method.

\begin{table}[]
    \centering
    \caption{\textbf{Effect of training different components:} We perform a quantitative study on the effect of updating or fixing different components of the \modelname architecture during training. We observe that the optimal strategy is to train the entire Reference U-Net and only train the RSA layers of the Base U-Net. Refer to Section \ref{subsec:ablation} for more details.  }
    \label{tab:ablation}
    \resizebox{0.97\linewidth}{!}{
        \begin{tabular}{c @{\hspace{0pt}} c c c ccc }
            \toprule
                   & \multicolumn{2}{c}{Trainable Parameters} & \multirow{2}[3]{*}{Use Aug.} & \multirow{2}[3]{*}{CLIP-T} & \multirow{2}[3]{*}{CLIP-I} & \multirow{2}[3]{*}{DINO}        \\
            \cmidrule{2-3}
                   & Base UNet                                & Ref. UNet                                                                                                                  \\
            \midrule
            \midrule
                   & All                                      & All                          & \cmark                     & \multicolumn{3}{c}{OOM}                                        \\
                   & None                                     & All                          & \cmark                     & 30.6                       & 72.8                       & 43.8 \\
                   & All                                      & None                         & \cmark                     & 31.6                       & 78.8                       & 66.4 \\
                   & RSA                                      & None                         & \cmark                     & 30.9                       & 78.2                       & 64.9 \\
                   & RSA                                      & All                          & \xmark                     & 30.2                       & 78.8                       & 67.0 \\
            \midrule
            (Ours) & RSA                                      & All                          & \cmark                     & 31.1                       & 79.7                       & 67.4 \\
            \bottomrule
        \end{tabular}
    }
\end{table}

\begin{figure}
\centering
    \includegraphics[width=0.8\columnwidth]{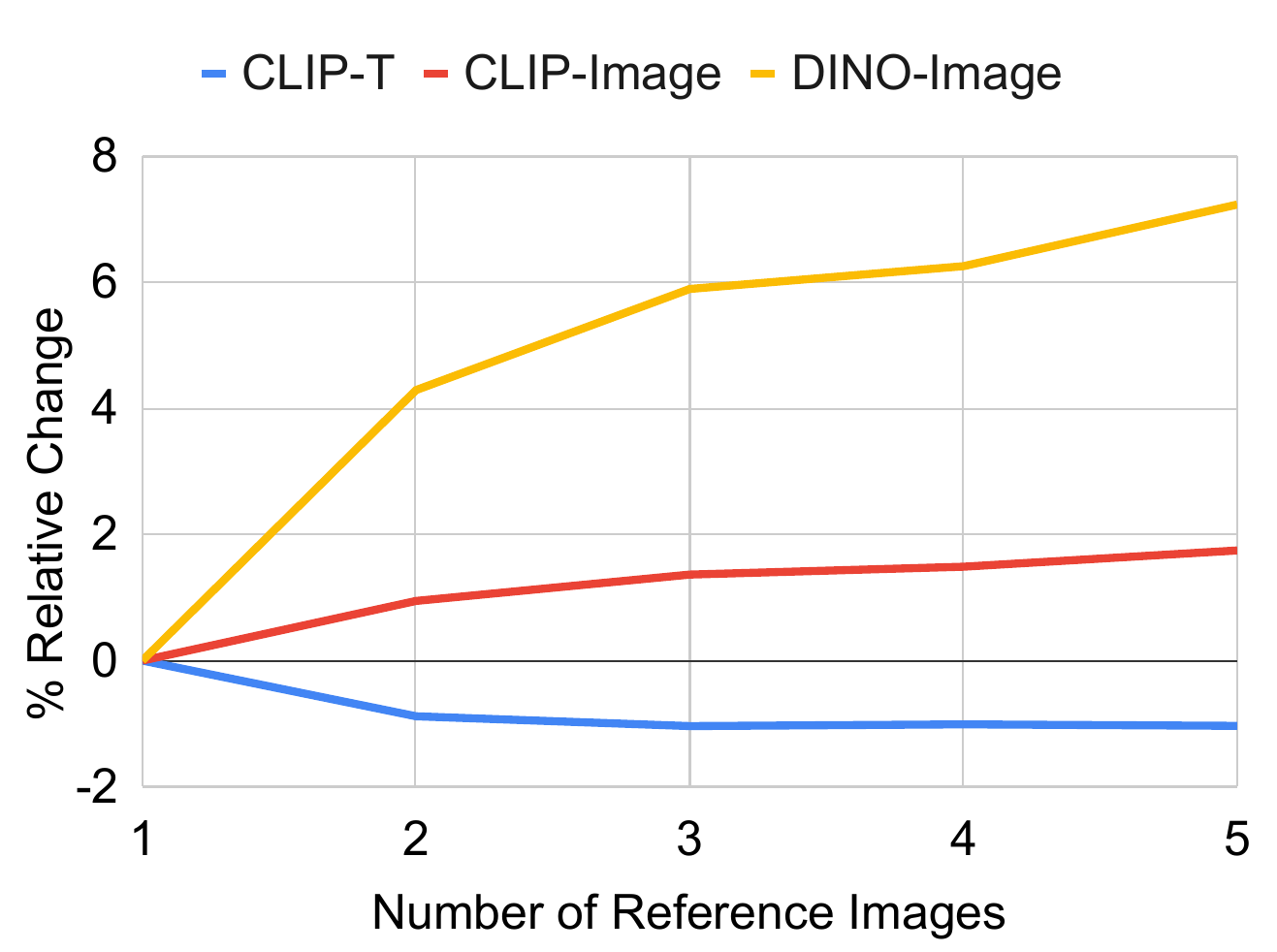}
    \caption{\textbf{Effect of multiple references}: We quantify the effect of using one or more reference images during inference. Increasing the number of reference images significantly improves subject fidelity ($+7.23$ DINO, $+1.74$ CLIP-I) while slightly worsening prompt fidelity ($-1.03$ CLIP-T)}
    \label{fig:num_ref}
\end{figure}

\subsection{Ablative Studies}
\label{subsec:ablation}

In order to understand the effect of different implementation choices, we further conduct ablative studies using the \modelname architecture. First, we study the effect of training/freezing different components of the \modelname architecture during the training process. In Table~\ref{tab:ablation}, we present the results on the three evaluation metrics. As discussed in Section~\ref{subsec:training}, for our best model, we train all the parameters of the Reference U-Net and only train the RSA layers of the Base U-Net. We find that the following a ControlNet style training approach, where only the Reference U-Net is trainable, leads to poor performance. We attribute this to the fact that the attention layers in the Base U-Net are not trained to handle features from multiple images. We also find that only training the RSA layers of the Base U-Net also leads to poor subject fidelity and text fidelity. This result highlight the importance of training the Reference U-Net to extract better reference features. Lastly, we find that finetuning both networks, is extremely memory intensive and hits our available memory limits even when training with small batch sizes. We also present results obtained after training our model without randomly augmenting the reference images (2nd last row). While we only see a minor drop in the subject fidelity metrics (CLIP-I, DINO), we observe in qualitative visualizations that the generated images are extremely corrupted (see supplementary material).

Next, we study the effect of the number of references used as input to \modelname. Since each reference image provides more information about the subject's appearance, the subject fidelity of an ideal personalized generation model should improve as the number of references increase. In Figure~\ref{fig:num_ref}, we present the relative change in the three metrics compared to outputs generated using only one reference image. We observe that the our subject fidelity metrics (CLIP-Image, DINO) consistently increase as the number of references increases. This highlights the efficacy of our proposed inference strategy that accounts for multiple reference images (introduced in Section~\ref{subsec:inference}).

Finally, we study different the efficacy of different inference strategies to handle multiple reference images with the \modelname architecture. In Figure~\ref{fig:refstrategy}, we present personalized generations on a few subjects using different inference strategies. ``Concat'' refers to concatenating the features from all reference images into one sequence before passing to the RSA layer. ``Average'' refers to averaging the outputs of an RSA layer after passing each reference image's feature in individually \textit{i.e.} $\frac{1}{K} \sum_{i=1}^K RSA(f, f_\texttt{refi})$. We observe that ``Concat'' often leads to generations where the object is repeated multiple times. This is expected since the model is receiving multiple copies of the same feature as input. On the other hand, the ``Average'' produces good quality images but smoothens out all the details in the objects. Our proposed inference strategy avoids these issues and generally produces higher quality personalized generations.

\section{Limitations}
\label{sec:limitations}
\modelname possesses many of the same failures as related methods. Specifically, in many instances, \modelname generation may fail to render fine-grained features of the subject and struggle to accurately adhere to the prompt. Some of these failure cases highlight the need to learn stronger fine-grained features, while other failure cases, such as prompt fidelity in certain conditions and text rendering (see Fig 4 row 5), are limitations inherited from the underlying text-to-image model. We provide further illustrations of failure cases in the Supplementary Material. Additionally, the weaknesses and the biases, including harmful stereotypes, of the underlying generative model will be perpetuated by \modelname generations. Subject-driven generation also opens the possibility of generating unwanted images of individuals without their consent. We urge that users of this technology are mindful of these implications and recommend that they use such technology responsibly.

\section{Discussion}
\label{sec:conclusion}
In this paper, we presented a bootstrapped training procedure that enables a text-to-image model to synthesize user-provided subjects in novel scenes \emph{without subject-specific finetuning}. Our method, \modelname, does not require real data or additional pretrained image encoders. Instead, it uses images generated by the text-to-image models, and utilizes a copy of the generative backbone to capture the appearance of user-provided subjects. The proposed model can be trained in approximately 1 hour, outperforms similar zero-shot inference methods and performs comparably to test-time finetuning methods. We believe that bootstrapped training of pretrained text-to-image models can be a promising paradigm for learning new capabilities and unlocking other modalities of controlled image generation.

{
    \small
    \bibliographystyle{ieeenat_fullname}
    \bibliography{main}
}

\clearpage
\onecolumn

\appendix

\section{Additional Details: Data Generation Pipeline}
\label{sec:supp_data}
As described in the main text, we utilize state-of-the-art pretrained models to synthesize training data. First, we use ChatGPT to generate captions for potential images. We use the following prompt for generating captions:
\begin{center}
    \resizebox{0.99\columnwidth}{!}{
        \begin{tcolorbox}
            Generate a caption for an image where there is one main object and possibly other secondary objects. The object needs to be finite and solid. The main object should not be things like lakes, fields, sky, etc.\vspace{10pt}

            You should only respond in the following format:
            Caption: A photo of a [object], [describe object positioning in scene] [describe scene]
        \end{tcolorbox}
    }
\end{center}

Some examples of the generated captions are:
\begin{center}
    \resizebox{0.99\columnwidth}{!}{
        \begin{tcolorbox}
            \begin{itemize}
                \item A photo of a red rose, nestled alone in a glass vase upon a lace table runner.
                \item A photo of a vintage typewriter, resting on a weathered desk.
                \item A photo of a faded Polaroid camera, lying on a sun-warmed picnic blanket.
                \item A photo of a sewing machine, buzzing with creativity amidst colorful fabric swatches in a designer's studio.
                \item A photo of a delicate porcelain teacup, delicately placed on a lace doily.
            \end{itemize}

        \end{tcolorbox}
    }
\end{center}

For each of these captions, we generate an image using Stable Diffusion 2.1\cite{rombach2022high} which is used as the target image. The formatting of the captions allows us to automatically extract the object categories by parsing the \texttt{[object]} part of the caption. For example, for the above captions, the corresponding objects categories are: \texttt{red rose, vintage typewriter, faded Polaroid camera, sewing machine, delicate porcelain teacup}. These object categories are used as input the Segment Anything Model\cite{Kirillov2023SegmentA} to extract foreground images. These foreground images pasted on a white background are used as reference images.

\clearpage
\section{Additional Details: Training and Inference}
\label{sec:supp_training}

In this section, we present additional implementation details for training and inference with the \modelname architecture. The source code to reproduce our experiments will be released with the next update to the paper.

\noindent
\textbf{Training} ~~~ In Algorithm~\ref{alg:bootpig}, we present a pseudocode describing the training pipeline. For optimizing the parameters, we use the AdamW algorithm with learning rate 5e-6, betas (0.9, 0.999), weight decay 1e-2 and epsilon 1e-8. The norm of the gradients is clipped to $1.0$.
\begin{algorithm}[h!]
    \caption{Training algorithm for \modelname.}\label{alg:bootpig}
    \begin{lstlisting}[mathescape=true] 
    Initialize: 
        $U_\theta, U_\phi$ = Stable Diffusion UNet
        VAE = Stable Diffusion VAE
        noise_scheduler = Stable Diffusion Noise Scheduler
        $U^{\texttt{RSA}}_\theta$ = Replace SA with RSA in $U_\theta$
        dataset = synthetic dataset containing 200k reference, target images and captions
        optimizer = AdamW( (RSA layers of $U_\theta$) + (All layers of $U_\phi$)   )

    def collect_features(model, latents, timestep, caption):
        Perform forward pass model(latents, timestep, caption)
        features = Collected features before each SA layer
        return features

    count = 0
    for reference, target, caption  in dataset:
        ref = VAE(reference)
        tar = VAE(target)
        t = random.randint(0, 1000)

        noise = randn_like(ref)
        noisedref = noise_sceduler.add_noise(ref, noise, t)
        noise = randn_like(tar)
        noisedtar = noise_sceduler.add_noise(tar, noise, t)

        $F_r = $ collect_features($U_\phi$, noisedref, t, caption)
        if random.rand()<0.15:
            caption = ""
        
        predtar = $U_\theta^{\texttt{RSA}}($noisedtar, t, caption, $F_r$)   # Use corresponding features from F_r in RSA layers 

        loss = ((noisedtar - predtar)**2).mean()
        loss.backward()
        count += 1
        if count %
            optimizer.step()
            optimizer.zero_grad()
    \end{lstlisting}
\end{algorithm}

\noindent
\textbf{Inference} ~~~ During inference, we first perform foreground segmentation on each reference image using the TRACER\cite{lee2022tracer} model implemented in the \texttt{carvekit} python package. Then we use the Reference UNet in the \modelname architecture to collect reference features for each reference image. We then perform denoising using the Base UNet with the pooling strategy described in Section 3.5.

\clearpage
\section{Additional Comparisons to Test-Time Finetuning Methods}
\label{sec:supp_compare}

\begin{figure*}[h!]
    \centering
    \includegraphics[width=0.95\textwidth]{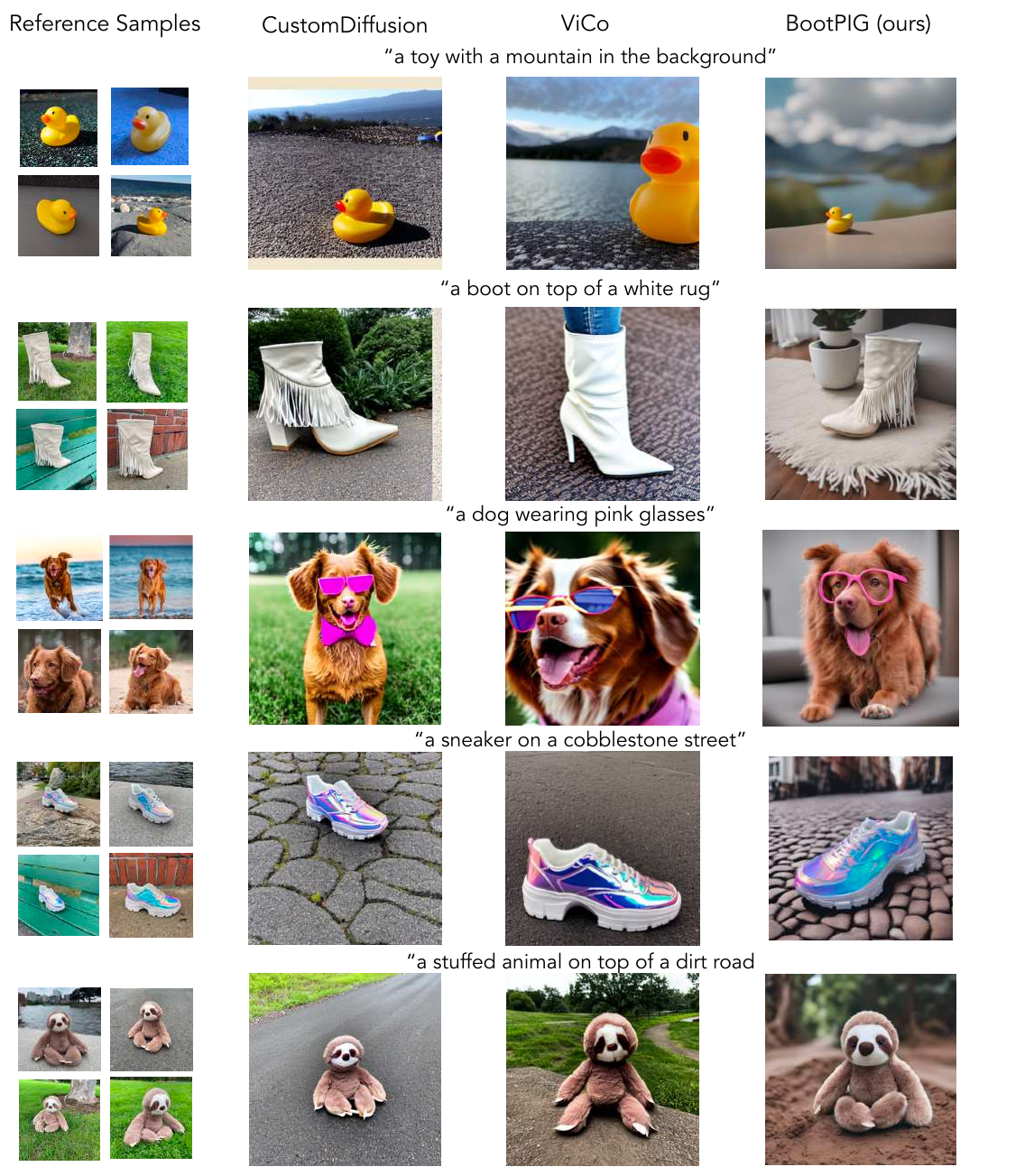}
    \caption{\textbf{Qualitative Comparision}: We provide additional visual comparisons of Custom Diffusion (reproduced), ViCo (reproduced), and \modelname. In comparision, \modelname generations demonstrate higher prompt fidelity (rows 2-4) and higher subject fidelity (rows 2, 3, 5).}
    \label{fig:supp_qualitative-comparision}
\end{figure*}

We compare our proposed method, \modelname, with two recent test-time finetuning approaches: CustomDiffusion \cite{kumari2023multi} and ViCo \cite{hao2023vico}. For both approaches, we train subject-specific models using each method on the DreamBooth \cite{ruiz2023dreambooth} dataset. We use the \texttt{diffusers} \cite{von_Platen_Diffusers_State-of-the-art_diffusion} implementation of CustomDiffusion and the official code provided by the authors of ViCo. We find that quantitatively (Table~\ref{tab:supp_quant}), \modelname outperforms CustomDiffusion ($+2.2$ CLIP-T, $+1.9$ DINO) and ViCo ($+3.1$ CLIP-T, and $+2.0$ CLIP-I \& $+8.9$ DINO) despite being zero-shot. These results highlight the benefits of \modelname as the method outperforms existing methods in prompt and subject fidelity, both quantitatively (CLIP-T, DINO) and qualitatively (Figure \ref{fig:supp_qualitative-comparision}).

We present additional qualitative comparisions with CustomDiffusion and ViCo on the DreamBooth dataset. Qualitative results, provided in Figure \ref{fig:supp_qualitative-comparision}, demonstrate the improvements in subject fidelity and prompt fidelity when using \modelname. In comparision to CustomDiffusion, we find that \modelname provides exhibits greater fidelity to the prompt. Examples of this improvement include: in row 2, CustomDiffusion's generation is missing the white rug; in row 3, CustomDiffusion adds an unnecessary pink bowtie to the subject; and in row 5 CustomDiffusion fails to place the stuffed animal on a dirt road. We also find noticeable improvements in subject fidelity when comparing \modelname to CustomDiffusion (shape of the duck in row 1, the length of the fringes on the boot in row 2, and color and thickness of dog's fur in row 3). Similarly, \modelname visually outperforms ViCo (\eg fails to match the details of the subject in rows 2, 3, 5 and does not follow the prompt in row 4).

\begin{table}[h!]
    \centering
    \begin{tabular}{@{\extracolsep{\fill}}lcccc@{}}
        \toprule
        Method                                             & Zero-shot & CLIP-T        & CLIP-I        & DINO          \\
        \midrule
        \midrule
        Ours                                               & \cmark    & \textbf{31.1} & \textbf{79.7} & \textbf{67.4} \\
        \midrule
        DreamBooth \cite{ruiz2023dreambooth}               & \xmark    & 30.5          & 80.3          & 66.8          \\
        CustomDiffusion\cite{kumari2023multi} (reproduced) & \xmark    & 28.9          & 80.6          & 65.5          \\
        ViCo \cite{hao2023vico} (reproduced)               & \xmark    & 28.0          & 77.7          & 58.5          \\
        \midrule
        Reference Images                                   & -         & -             & 88.5          & 77.4          \\
        \bottomrule
    \end{tabular}
    \caption{\textbf{Quantitative Comparisons:} We provide average metrics for subject fidelity (CLIP-I, DINO) and prompt fidelity (CLIP-T) on the DreamBooth dataset.}
    \label{tab:supp_quant}
\end{table}

\clearpage
\section{Additional Qualitative Results}
\label{sec:supp_qual}
\begin{figure*}[h!]
    \centering
    \includegraphics[width=0.85\textwidth]{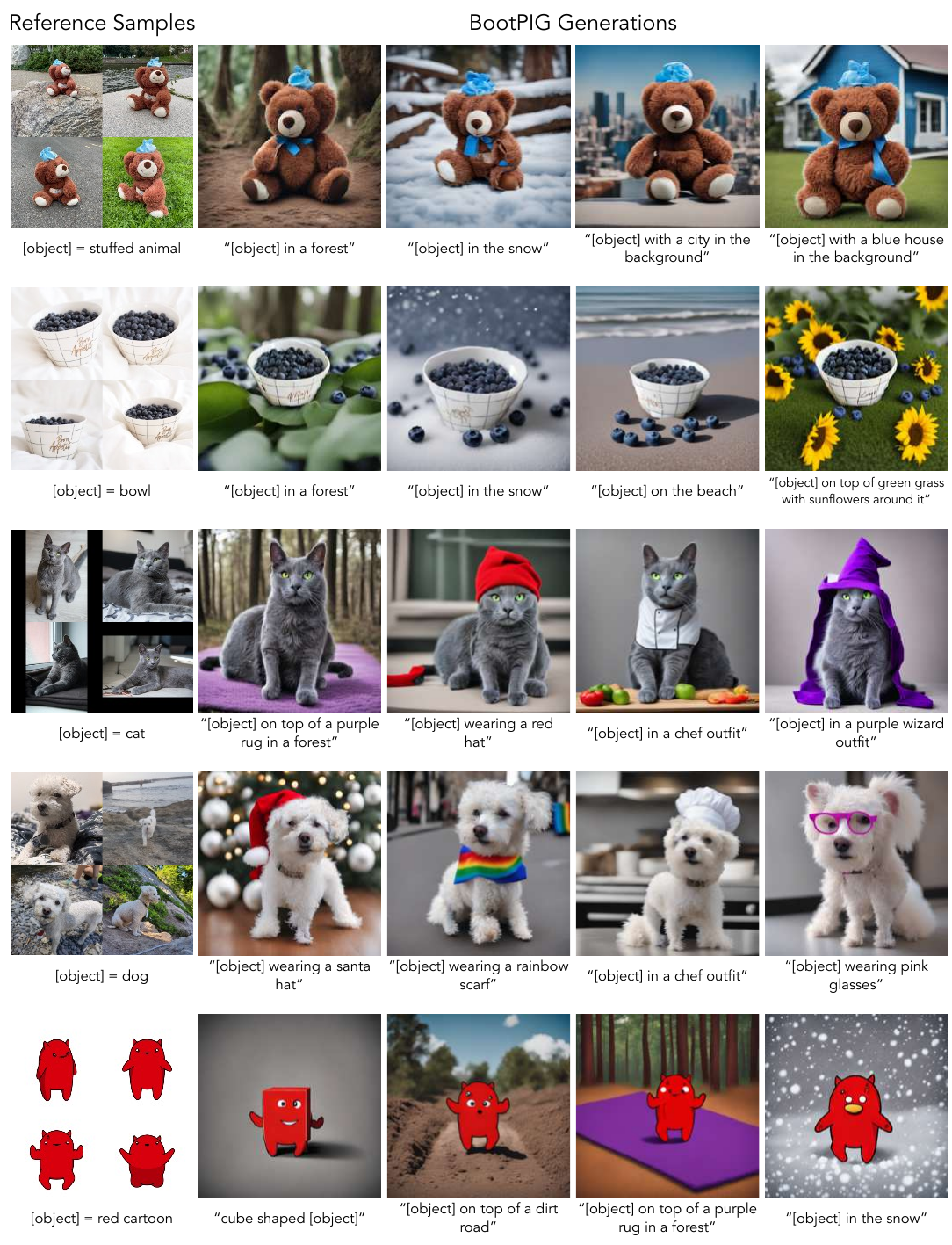}
    \vspace{-2pt}
    \caption{\textbf{Additional Visualizations:} We present additional qualitative results on the DreamBooth dataset using our proposed method.}
    \label{fig:supp_visualizations}
\end{figure*}

In Figure~\ref{fig:supp_visualizations}, we present additional visualizations using our proposed method on several DreamBooth dataset images. We observe that across different subjects, our proposed method is able to successfully maintain subject fidelity while accurately following input prompts.

\clearpage
\section{Additional Qualitative Ablation}
\label{sec:supp_initrefaug}

\begin{figure*}[b!]
    \centering
    \includegraphics[width=0.8\textwidth]{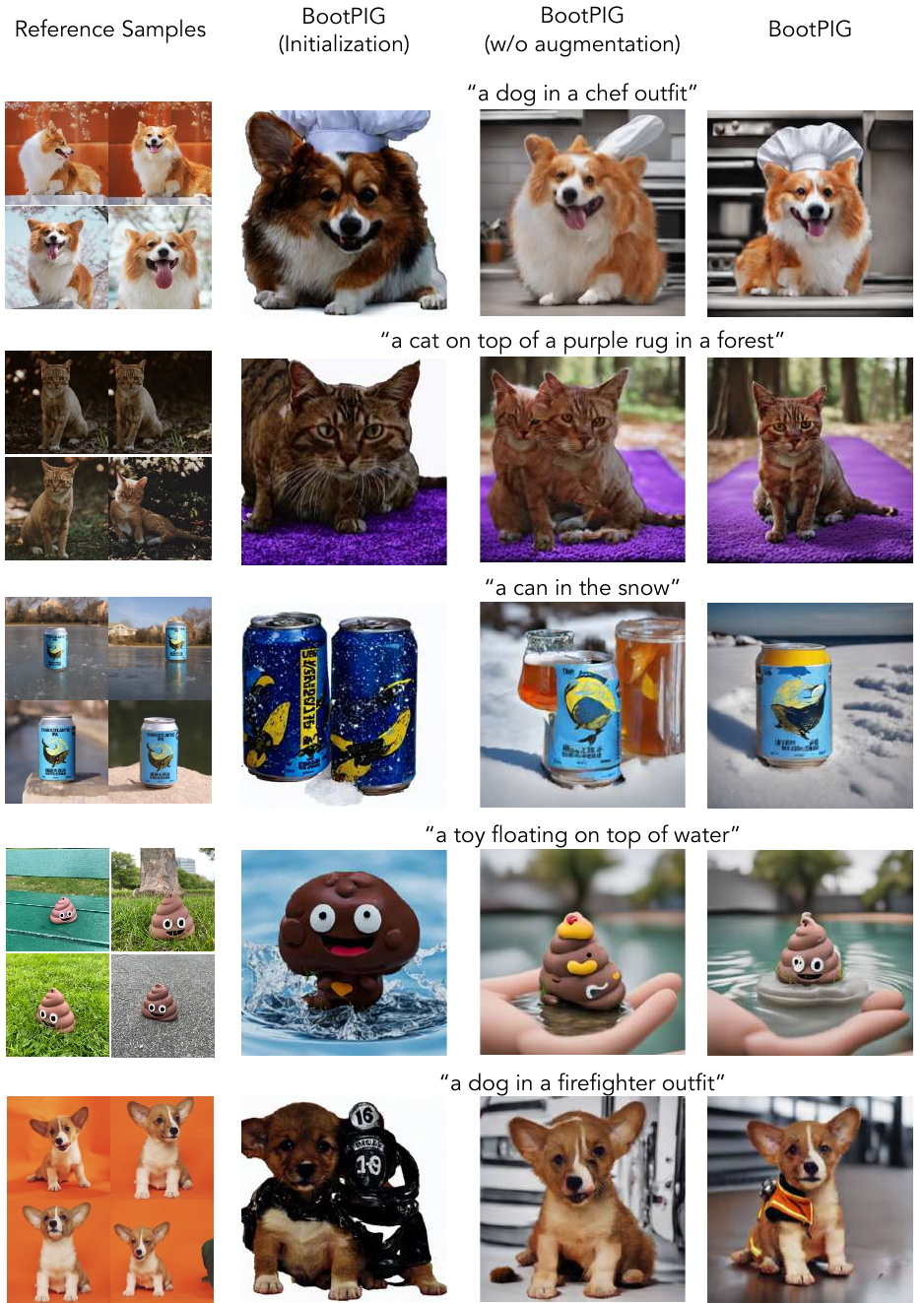}
    \caption{\textbf{Qualitative Ablation:}  }
    \label{fig:supp_initvis}
\end{figure*}

In Figure~\ref{fig:supp_initvis}, we present additional qualitative results to ablate the effect of training the \modelname architecture and the importance of augmenting the reference images during training. At initialization, we observe that the \modelname architecture can already copy some aspects of the subject's appearance. However, both the subject and prompt fidelity is limited, and the overall quality of generated images is poor.

\noindent
In Section 4.4 of the main text, we quantitatively show that synthetically augmenting the reference images leads to improvements in final performance. Specifically, we random horizontal flipping, randomly resize the foreground in the images and place the foreground at random locations in the white background images. We observed that the subject fidelity metrics (CLIP-I, DINO-I) demonstrate minor improvements when \modelname is trained with such augmented images. However, in practice, we observed that training without augmentations led to significantly worse subject preservation and overall image quality. In Figure~\ref{fig:supp_initvis} Column 2 \& 3, we demonstrate this by qualitatively comparing the effect of the synthetic augmentations.

\clearpage

\section{Subject-Driven Inpainting}
\label{subsec:inpainting}
\begin{figure}[htb]
\centering
    \includegraphics[width=0.75\columnwidth]{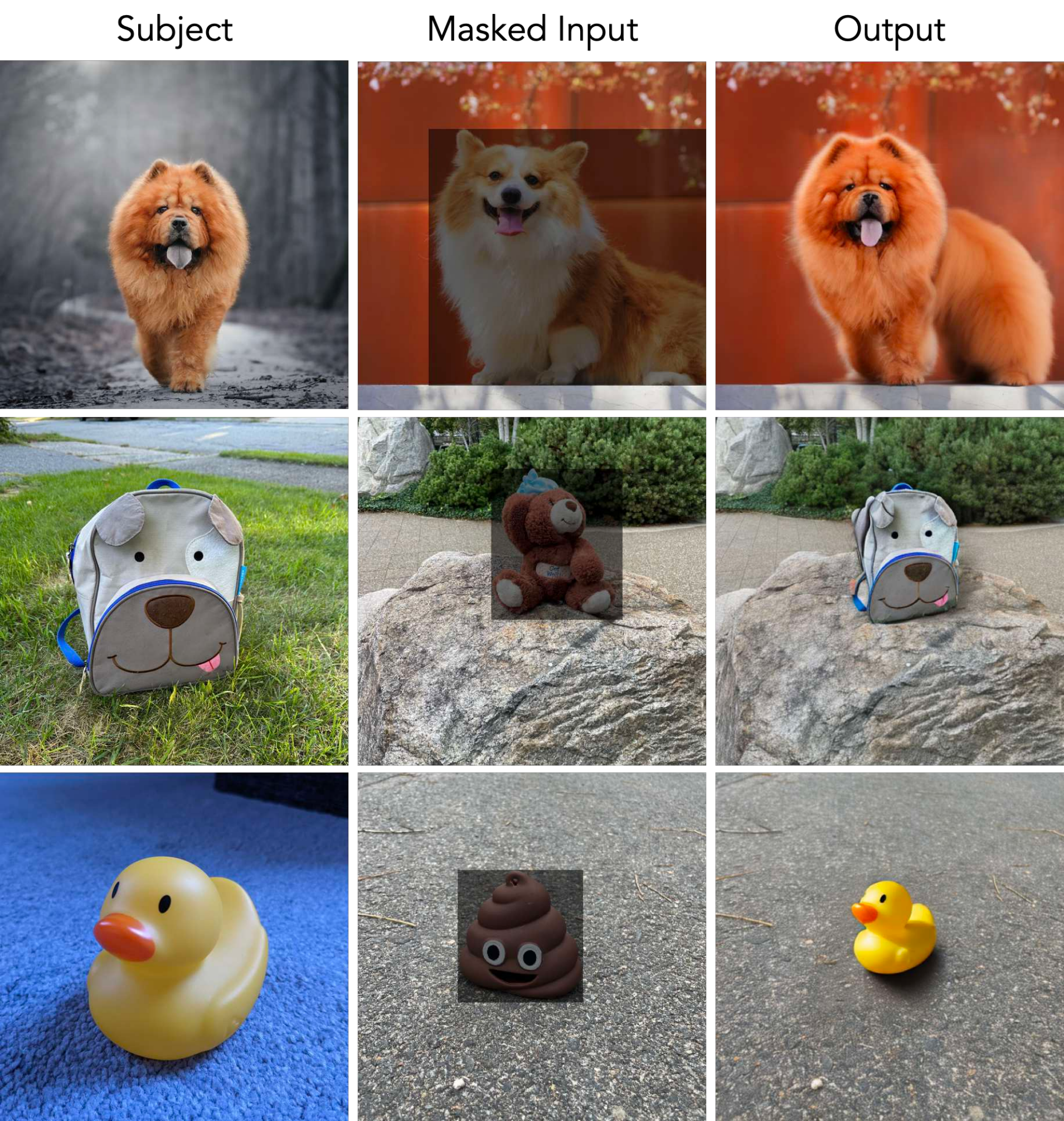}
    \caption{\textbf{Zero-shot Subject-Driven Inpainting}: We provide visual comparisons of \modelname inpainted images according to a given subject. Despite not being trained using an inpainting objective, we find that \modelname is able to accurately render the subject within the original scene. }
    \label{fig:inpaint}
\end{figure}

We additionally qualiatively explore the zero-shot capabilities of \modelname for subject-driven inpainting. Given an input image, prompt, and binary segmentation mask, we use \modelname to inpaint a given subject into the masked region of the image. We provide examples of \modelname inpainted images in Figure \ref{fig:inpaint}. Similar to the results in the text-to-image experiments, we find that \modelname can preserve the details of the target subject in the new scene (\eg face and fur of the dog in row 1, the dog face on the backpack in row 2, and the face and wing of the duck in row 3). We note that \modelname was not trained using an inpainting objective, thus limiting its capabilities as an image editing model. As a result, we find that \modelname can struggle to inpaint target subjects when they are drastically different, in terms of size and/or shape, to the existing subject in the scene.

\clearpage
\section{Failures Cases}
\label{sec:supp_failure}

\begin{figure*}[h!]
    \centering
    \includegraphics[width=0.9\textwidth]{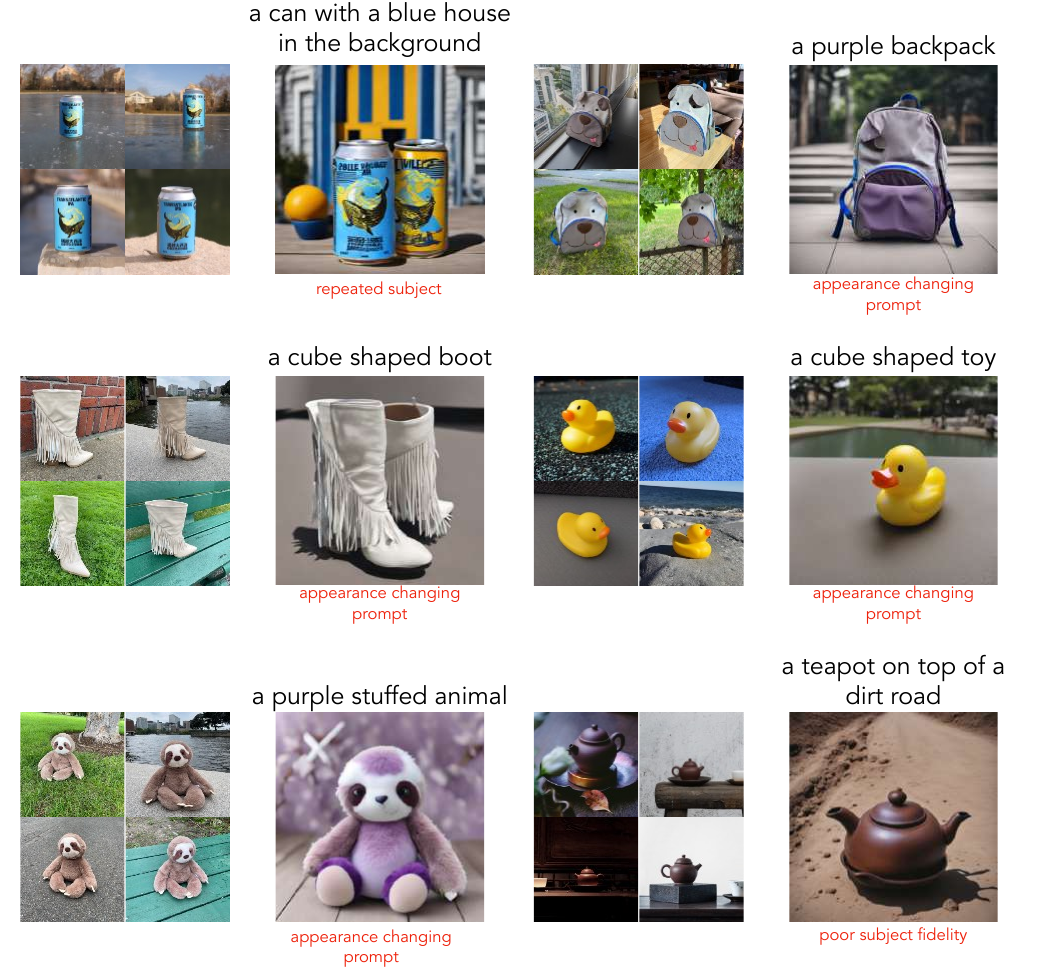}
    \caption{\textbf{Failure Cases:} We present examples of a few failure modes demonstrated by the \modelname architecture.}
    \label{fig:supp_failure}
\end{figure*}

The \modelname architecture does occasionally fail on certain (subject, caption) pairs. In Figure~\ref{fig:supp_failure}, we present examples of some failed generations. We observe that a commonly occurring failure is in the case where the prompt attempts to modify the appearance of the subject. For example, prompts of the form "a cube shaped \texttt{[object]}" attempts to modify the shape of reference \texttt{[object]} to a cube. Since the \modelname architecture was trained on synthetic data where the target and reference appearance accurately matches, it fails to capture such modifications to the appearance during inference. We also observed that the \modelname architecture occasionally generates images depicting multiple instances of the reference object. Furthermore, as with any existing personalization method, our proposed model also occasionally fails at accurately capturing the appearance of the object (see Figure~\ref{fig:supp_failure} row 3, column 2). As a guideline for practitioners, we find that regenerating the image with a different sample of Gaussian noise often resolves the issue.

\end{document}